\documentclass[sigconf]{acmart}

\AtBeginDocument{%
  }

\setcopyright{acmlicensed}
\copyrightyear{2024}
\acmYear{2024}
\acmDOI{XXXXXXX.XXXXXXX}

\acmConference[CIKM '24]{33rd ACM International Conference on
Information and Knowledge Management}{October 21--25,
  2024}{Boise, ID, USA}
\acmISBN{978-1-4503-XXXX-X/18/06}

\usepackage{url}            
\usepackage{booktabs}       
\usepackage{amsfonts}       
\usepackage{nicefrac}       
\usepackage{microtype}      
\usepackage{amsmath}
\usepackage{courier, paralist}
\usepackage{bm}
\usepackage{algorithm}
\usepackage{algorithmic}
\usepackage{color}
\usepackage{enumitem}
\usepackage{graphicx}
\usepackage{array}
\usepackage{multirow}
\usepackage{setspace}
\usepackage{subcaption}
\usepackage{makecell}
\usepackage{balance}
\usepackage{mathtools}
\usepackage{tikz}
\usetikzlibrary{shapes.geometric}
\usetikzlibrary{arrows.meta,arrows}
\usepackage{tabularx}
\usepackage{xspace}
\usepackage{stfloats}
\usepackage[hang,flushmargin,multiple]{footmisc}
\definecolor{OliveGreen}{cmyk}{0.64,0,0.95,0.40}

\newcommand{\stt}{\mathbf{s}_t}
\newcommand{\stp}{\mathbf{s}_{t+1}}

\newcommand{\at}{\mathbf{a}_t}

\newcommand{\bs}{\mathbf{s}}
\newcommand{\ba}{\mathbf{a}}

\newcommand{\btau}[1]{\bm{\tau}^{#1}}

\newcommand{\expect}[2]{\mathbb{E}_{#1} \left[ #2 \right] }

\newcommand{\te}[1]{\texttt{#1}}
\DeclareMathOperator*{\argmax}{arg\,max}

\def\vf{{\bm{f}}}

\def\vs{{\bm{s}}}

\newcommand\blfootnote[1]{%
  \begingroup
  \renewcommand\thefootnote{}\footnote{#1}%
  \addtocounter{footnote}{-1}%
  \endgroup
}

\newcommand{\ours}{\textsc{PlayBest}\xspace}

\begin{document}

\title{\ours: Professional Basketball Player Behavior Synthesis \\via Planning with Diffusion}

\author{Xiusi~Chen$^{*}$}
\affiliation{%
  \institution{University of California, Los Angeles}
  \city{}
  \state{}
  \country{}
}
\email{xchen@cs.ucla.edu}

\author{Wei-Yao~Wang$^{*}$}
\affiliation{%
  \institution{National Yang Ming Chiao Tung University}
  \city{}
  \state{}
  \country{}
}
\email{sf1638.cs05@nctu.edu.tw}

\author{Ziniu~Hu}
\affiliation{%
  \institution{California Institute of Technology}
  \city{}
  \state{}
  \country{}
}
\email{acgbull@gmail.com}

\author{David~Reynoso}
\affiliation{%
  \institution{University of California, Los Angeles}
  \city{}
  \state{}
  \country{}
}
\email{dmreynos@g.ucla.edu}

\author{Kun~Jin}
\affiliation{%
  \institution{University of Michigan, Ann Arbor}
  \city{}
  \state{}
  \country{}
}
\email{kunj@umich.edu}

\author{Mingyan~Liu}
\affiliation{%
  \institution{University of Michigan, Ann Arbor}
  \city{}
  \state{}
  \country{}
}
\email{mingyan@umich.edu}

\author{P.~Jeffrey~Brantingham}
\affiliation{%
  \institution{University of California, Los Angeles}
  \city{}
  \state{}
  \country{}
}
\email{branting@g.ucla.edu}

\author{Wei~Wang}
\affiliation{%
  \institution{University of California, Los Angeles}
  \city{}
  \state{}
  \country{}
}
\email{weiwang@cs.ucla.edu}

\renewcommand{\shortauthors}{Chen et al.}

\begin{abstract}
Dynamically planning in complex systems has been explored to improve decision-making in various domains.
Professional basketball serves as a compelling example of a dynamic spatio-temporal game, encompassing context-dependent decision-making.
However, processing the diverse on-court signals and navigating the vast space of potential actions and outcomes make it difficult for existing approaches to swiftly identify optimal strategies in response to evolving circumstances. In this study, we formulate the sequential decision-making process as a conditional trajectory generation process. Based on the formulation, we introduce \ours (PLAYer BEhavior SynThesis), a method to improve player decision-making. We extend the diffusion probabilistic model to learn challenging environmental dynamics from historical National Basketball Association (NBA) player motion tracking data. To incorporate data-driven strategies, an auxiliary value function is trained with corresponding rewards. To accomplish reward-guided trajectory generation, we condition the diffusion model on the value function via classifier-guided sampling. We validate the effectiveness of \ours through simulation studies, contrasting the generated trajectories with those employed by professional basketball teams. Our results reveal that the model excels at generating reasonable basketball trajectories that produce efficient plays. Moreover, the synthesized play strategies exhibit an alignment with professional tactics, highlighting the model's capacity to capture the intricate dynamics of basketball games.\footnote{The code is at \url{https://github.com/xiusic/diffuser_bball}.}
\end{abstract}

\ccsdesc[500]{Computing methodologies~Planning for deterministic actions}
\ccsdesc[500]{Computing methodologies~Machine learning}

\keywords{Planning, Diffusion model, Conditional sampling, Sports analytics}


\maketitle

\blfootnote{$^*$Equal contribution.}
\section{Introduction}
\label{sec:intro}
The exploration of dynamic systems and their planning has broad applicability across various domains. Whether it involves developing strategies for team sports \citep{wang2018advantage}, managing traffic flow \citep{zang2020metalight}, coordinating autonomous vehicles \citep{kiran2021deep}, or understanding the dynamics of financial markets \citep{liu2022finrl}, these scenarios can be effectively framed as dynamic systems characterized by intricate interactions and decision-making processes. The ability to comprehend and plan within these systems becomes crucial to achieving optimal outcomes. Basketball, with its high level of dynamism and complexity as a team sport, serves as a captivating illustration of a real-time dynamic system with intricate tactical elements.
A basketball game requires continuous adaptation and strategic decision-making. Coaches and players rely on pertinent environmental and behavioral cues including teammates' and opponents' current positions and trajectories to select play strategies that respond effectively to opponents' actions and adapt to real-time situational changes. Existing methods in sports analytics and trajectory optimization \citep{wang2018advantage,terner2020modeling,wang2022shuttlenet} have made progress in modeling and predicting player movements and game outcomes. However, these approaches struggle to capture the intricate dynamics of basketball games and produce flexible, adaptive play strategies that can handle the uncertainties and complexities inherent in the sport. The challenges arise from the following two features of basketball games:


\noindent\textbf{Modeling the complex environmental dynamics:} Capturing environmental dynamics in basketball games is a very challenging task due to the inherent complexity of the game, for example, rapid changes in game situations and numerous possible actions at any given moment.
The spatio-temporal nature of basketball data, including multiple player positions and ball trajectories, further complicates the modeling process. The need for a computationally efficient and scalable approach to handle the massive amounts of data generated during basketball games presents a major challenge in modeling environmental dynamics.

\noindent\textbf{Reward Sparsity:} An additional challenge lies in addressing reward sparsity. Unlike other reinforcement learning (RL) environments where immediate feedback is readily available after each action, basketball games often see long sequences of actions leading up to a single reward event (e.g., the scoring of a basket). This results in a sparse reward landscape, as many actions contribute indirectly to the final outcome but are not themselves immediately rewarded. This scenario complicates the learning process as it becomes more challenging for the planning algorithm to accurately attribute the impact of individual actions to the final reward. Designing effective methods to address the reward sparsity challenge remains a significant hurdle in applying typical planning algorithms to basketball and similar sports games.


Recently, powerful trajectory optimizers that leverage learned models often produce plans that resemble adversarial examples rather than optimal trajectories \citep{talvitie2014model,ke2019modeling}. On the contrary, modern model-based RL algorithms tend to draw more from model-free approaches, such as value functions and policy gradients \citep{wang2019benchmarking}, rather than utilizing the trajectory optimization toolbox. Methods that depend on online planning typically employ straightforward gradient-free trajectory optimization techniques like random shooting \citep{nagabandi2018neural} or the cross-entropy method \citep{botev2013cross,chua2018deep} to circumvent the above problems.

In this work, we first formulate the planning problem in basketball as a multi-player behavior synthesis task, and instantiate the behavior synthesis task as a trajectory generation task. Following the recent success of generative models in applications of single-agent planning \citep{janner2022planning,ajay2022conditional}, we propose a novel application of the diffusion model called \ours (PLAYer BEhavior SynThesis), to generate optimal basketball trajectories and synthesize adaptive play strategies. Under most circumstances, the diffusion model serves as a generative model to capture the distribution of the input samples. In our study, we extend it as a powerful technique to enable flexible behavior synthesis in dynamic and uncertain environments since there is no existing online environment for basketball simulations. The diffusion process explores different potential trajectories and adapts to changes in the environment through the iterative sampling process to model basketball court dynamics. To guide the reverse diffusion process with rewards, \ours features a value guidance module that guides the diffusion model to generate optimal play trajectories by conditional sampling. This integration naturally forms a conditional generative process, and it allows \ours to swiftly adapt to evolving conditions and pinpoint optimal solutions in real-time.

We instantiate \ours in a variety of simulation studies and real-world scenarios, demonstrating the effectiveness of \ours in generating high-quality basketball trajectories that yield effective plays. Extensive results reveal that our proposed approach outperforms conventional planning methods in terms of adaptability, flexibility, and overall performance, showing a remarkable alignment with professional basketball tactics.

The core contributions of this work are summarized as follows:
\begin{itemize}[nosep,leftmargin=*]  
    \item We attempt to formulate the basketball player behavior synthesis problem as a guided sampling/conditional generation of multiple players and ball trajectories from diffusion models.
    \item We present \ours, a framework featuring a diffusion probabilistic model with a value function, to instantiate the conditional generative model. We adapt the model to integrate multi-player behaviors and decisions in basketball and show that a number of desirable properties are obtained.
    \item We showcase the effectiveness of \ours via both quantitative and qualitative studies of the trajectories generated and validate the practicality of adopting \ours to investigate real basketball games.
\end{itemize}

\section{Preliminary}

\subsection{Diffusion Probabilistic Models}
\label{sec:background_diffusion}

Diffusion probabilistic models \citep{sohl2015deep,ho2020denoising} stand out as a unique approach to learning complex data distributions, symbolized by $q( \btau{} )$, based on a collection of samples, denoted as $\mathcal{D} \coloneqq \{\boldsymbol{x}\}$.

On a high level, two processes are at the core of their operation: a predefined forward noising mechanism $q(\btau{i+1} | \btau{i}) \coloneqq \mathcal{N}(\btau{i+1}; \sqrt{\alpha_{i}}\btau{i},\\ (1-\alpha_{i})\boldsymbol{I})$ and a trainable reverse or ``denoising'' process $p_{\theta}(\btau{i-1}|\btau{i}) \\ \coloneqq \mathcal{N}(\btau{i-1}|\mu_{\theta}(\btau{i}, i), \Sigma_{i})$.
Here the Gaussian distribution is represented as $\mathcal{N}(\mu, \Sigma)$, and the variable $\alpha_i$ is pivital in determining the variance schedule. We begin with a sample $\boldsymbol{x}_{0} \coloneqq \boldsymbol{x}$, followed by latents $\btau{1}, \btau{2}, ..., \btau{N-1}$, and culminate with $\btau{N} \sim \mathcal{N}(\boldsymbol{0}, \boldsymbol{I})$, factoring in specific values for $\alpha_{i}$ and an adequately extended $N$.

\subsection{Trajectory Optimization Problem Setting in Basketball Strategy}
\label{sec:background_traj_opt}
In basketball, we can consider the game as a discrete-time system with dynamics $\stp = \vf(\stt, \at)$, where $\stt$ represents the state of the play, and $\at$ denotes the action or basketball maneuver. Trajectory optimization aims to find a sequence of actions $\ba_{0:T}^*$ that maximizes an objective $\mathcal{J}$, such as maximizing the score. This can be represented as:
\begin{equation}
\ba_{0:T}^* = \argmax_{\ba_{0:T}} \mathcal{J}(\bs_0, \ba_{0:T}) = \argmax_{\ba_{0:T}} \sum_{t=0}^{T} r(\stt, \at)
\end{equation}
where $T$ defines the planning horizon. $\btau{} = (\bs_0, \ba_0, \bs_1, \ba_1, \ldots, \bs_T, \ba_T)$ is the trajectory of states and actions, and $\mathcal{J}$ becomes the objective value of the play.

This model, when applied to basketball, facilitates the creation of dynamic strategies that adapt to real-time game scenarios. By simulating noise-corrupted play sequences and iteratively denoising them, one can derive actionable insights into players' behaviors, leading to more effective in-game decision-making and planning.

\begin{figure*}[t!]
    \centering
    \includegraphics[width=.95\linewidth]{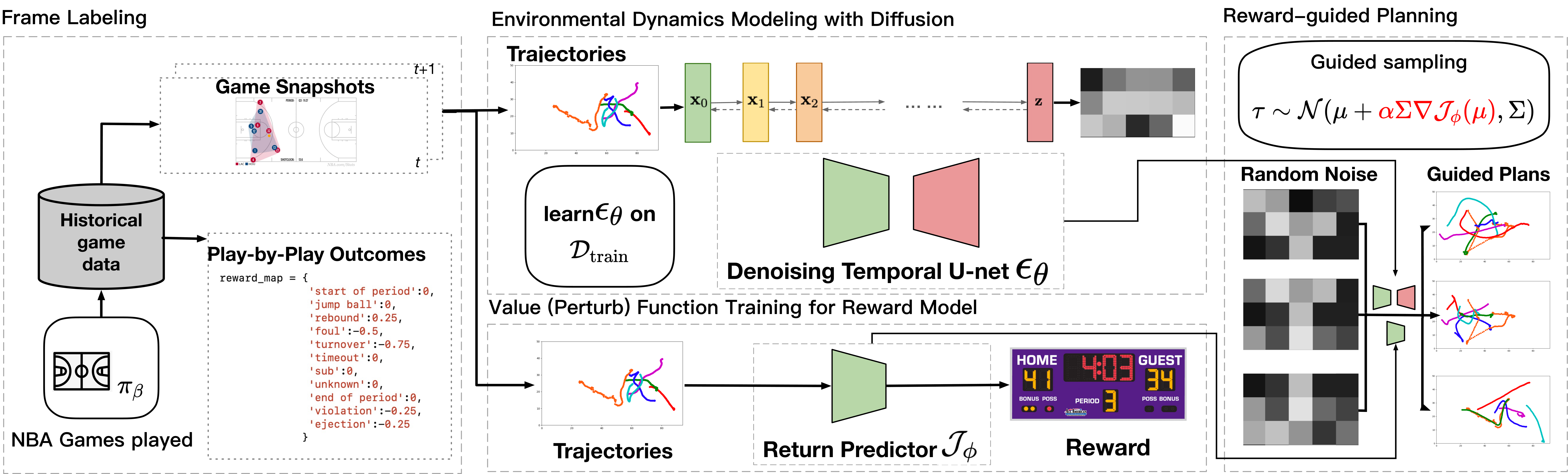}
    \caption{\textbf{Overview framework of \ours.} The overall pipeline can be split into four major components: Frame Labeling, Environmental Dynamics Learning, Value (Perturb) Function Training, and Trajectory Generation Guided by a Reward Function. The diffusion probabilistic model $\epsilon_\theta$ is trained to model the environmental dynamics. The reward predictor $\mathcal{J}_\phi$ is trained on the same trajectories as the diffusion model. During guided trajectory generation, our model takes both environmental dynamics and rewards as input, performs guided planning via conditional sampling, and generates the trajectories as the guided plan.}
    \label{fig:overview}
\end{figure*}


\subsection{Problem Description}

The input for \ours consists of a set of basketball game records, denoted as $\mathcal{D}_{raw}$. These game records are composed of distinct elements, described as follows:


\noindent \textbf{Motion Track Data.} The motion track data, represented as $\mathcal{D}^{move}$, comprises static snapshots of in-game events, detailing the positions of all players and the ball at a rate of 25 frames per second. A game's progression can be reconstructed and visualized using these snapshots.


\noindent \textbf{Play-by-Play Data.} Denoted as $\mathcal{D}^{pbp}$, the play-by-play data offers a game transcript in the form of possessions. This data includes 1) the possession timestamp, 2) the player initiating the possession, 3) the result of the possession (e.g., points scored), and 4) additional unique identifiers employed for possession categorization.




To facilitate learning, we divide $\mathcal{D}_{raw}$ into $\mathcal{D}_{train}$ and $\mathcal{D}_{test}$, representing the training and testing sets, based on gameplay timestamps. We formally define our task as follows:


Given a set of game records $\mathcal{D}_{train} = \mathcal{D}_{train}^{move} \cup \mathcal{D}_{train}^{pbp}$ and a reward function $\mathcal{J}_{\phi}$, with $\mathcal{J}_{\phi}$ depending on the reward definition given by the discriminative rules applied to $\mathcal{D}_{train}^{pbp}$, the objective is to generate trajectories $\{\btau{} \}$ leaning towards the higher-reward regions of the state-action space. In essence, our goal is to develop a policy $\pi_{\theta,\phi}(\mathbf{a} \mid \mathbf{s})$, parameterized by $\theta$ and $\phi$, that determines the optimal action based on the states associated with each frame in $\mathcal{D}_{test}^{move}$.

\section{The \ours Framework}
\label{sec:frame}
In this section, we describe in detail how our framework is designed. We first give an overview and then present details of the model architecture including the diffusion and value function modules.

\subsection{Framework Overview}
Figure~\ref{fig:overview} depicts the \ours pipeline. The historical game replay data originates from actual games played during the 2015-2016 NBA regular season. Each team competes per their unknown policies $\pi_\beta$. The raw game data encompasses multiple modalities, and a game is characterized by a series of high-frequency snapshots (e.g., 25 frames per second). At any given time $t$, a snapshot includes an image displaying all player and ball positions, as well as additional metadata like the results of each possession (shot made/miss, free-throw made/miss, rebound, foul, turnover, etc), shot clock, and game clock at time $t$. 


Out of the historical game replay data, we construct the player trajectories and ball trajectories to create the trajectory dataset $\mathcal{D}^{move}$. We then use the trajectory dataset $\mathcal{D}_{train}^{move}$ to train a diffusion model $\epsilon_\theta$ that aims at modeling the distribution of the 3-dimensional player and ball movements. The training process of the diffusion model mimics the training procedure of what is usually referred to as offline RL, where there is no online environment to interact with. However, the diffusion model by itself can only generate ``like-real'' trajectories that do not necessarily lead to a goal-specific outcome. To further generate trajectories that can represent ``good plans'', we train a value function that maps any possible trajectory to its expected return. During the sampling stage, the mean of the diffusion model is perturbed by the gradient of the value function. In this way, the guided sampling is capable of generating the trajectories biased towards the high-reward region. Incorporating a diffusion model in planning problems not only enhances efficient exploration and resilience in volatile environments, but also addresses the challenge of long-horizon planning, enabling the generation of strategic, noise-reduced trajectories over extended periods.

\begin{figure*}[t!] 
\centering
\begin{subfigure}{0.4\textwidth}
  \centering
  \includegraphics[width=.85\linewidth]{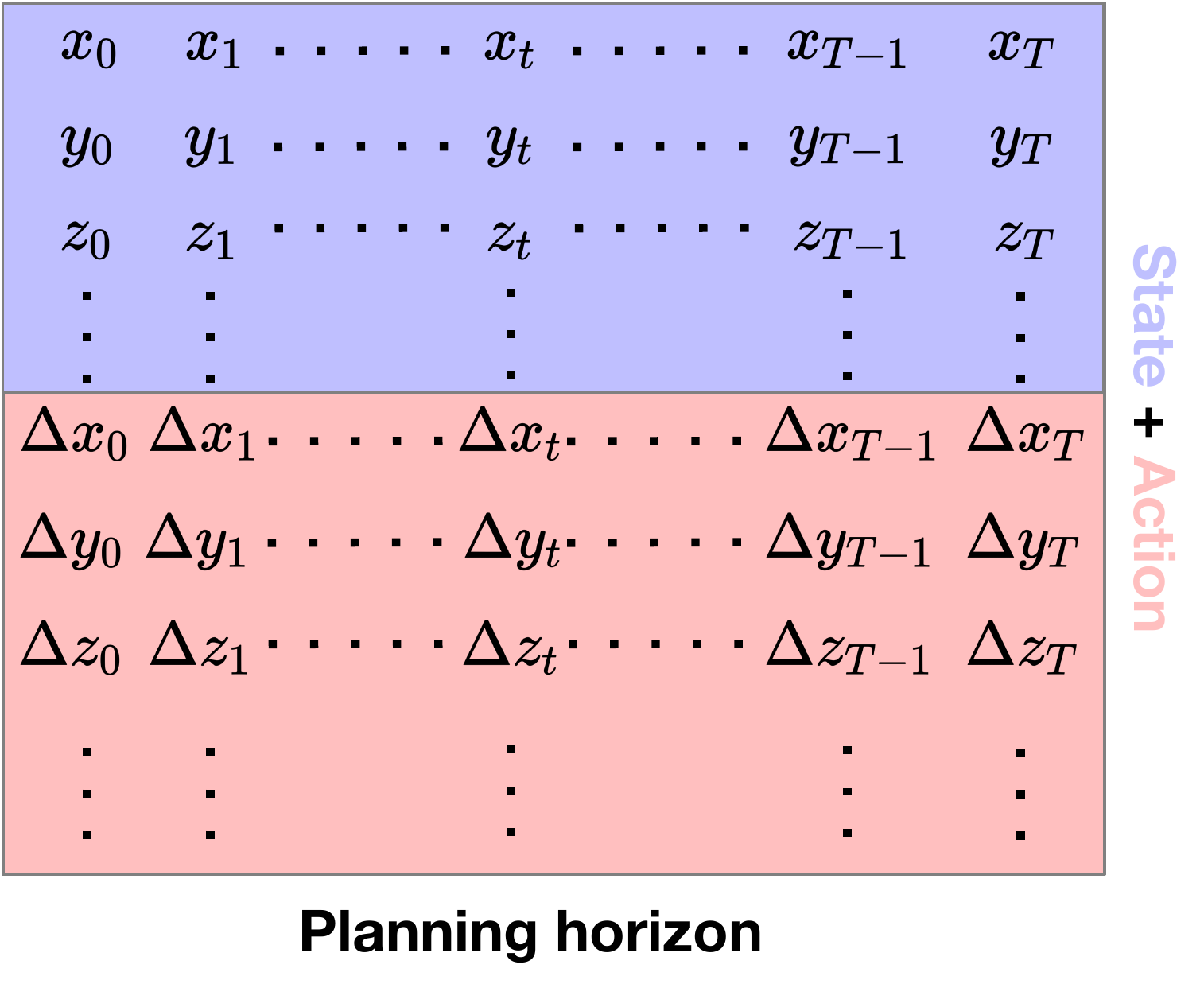}
  \caption{The shape of the training data. Trajectories are represented by the $(x, y, z)$ coordinates of the ten on-court players across two teams and the ball (11 channels). The action is made up of the momentum of each object at the same timestep.}
\label{fig:input}
\end{subfigure}
\ \ \ \ \
\begin{subfigure}{0.5\textwidth}
  \centering
  \includegraphics[width=\linewidth]{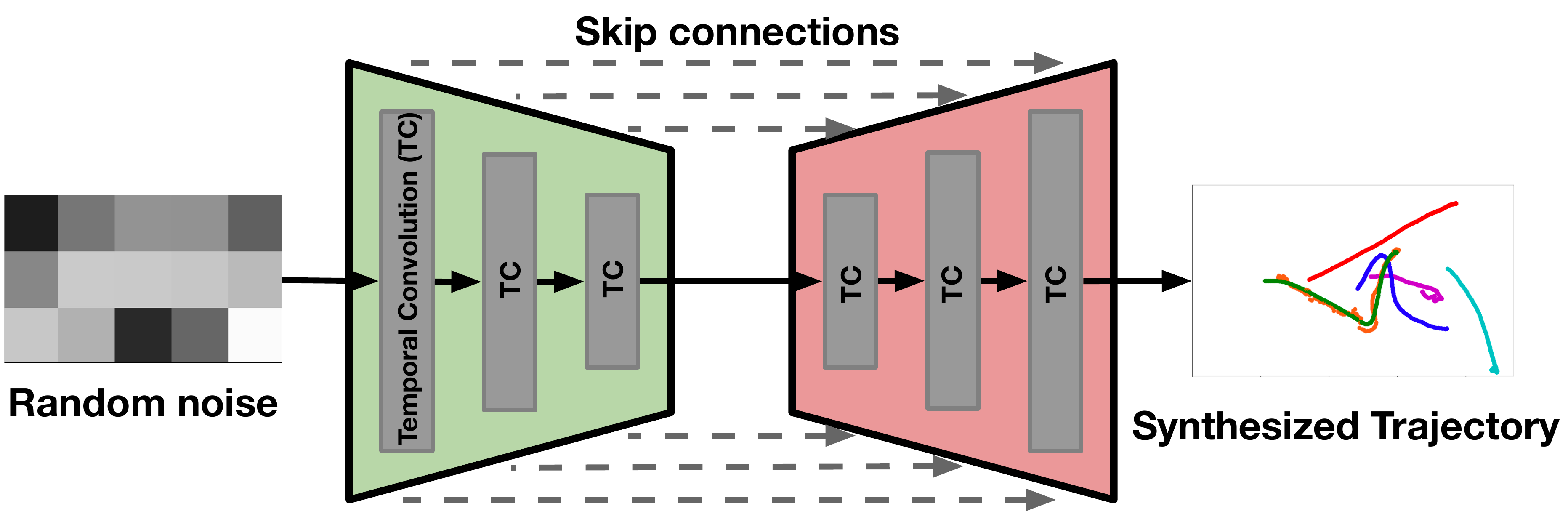}
  \vspace{15pt}
  \caption{The general structure of the diffusion model $\epsilon_\theta$ is implemented by a U-net with temporal convolutional blocks, which have been widely utilized in image-centric diffusion models.}
  \label{fig:architect}
\end{subfigure}
\caption{\textbf{(a, b)} \textbf{The input and diffusion architecture.}}
\label{fig:input_archi}
\end{figure*}

In essence, our framework utilizes a dataset $\mathcal{D}$ collected by an unknown behavior policy $\pi_\beta$, which can be approximated as the ``average" policy for all NBA teams. This dataset is gathered once and remains unaltered during training. The training process relies entirely on the training set $\mathcal{D}_{\text {train }}$ and does not interact with the environment. Upon completion of training, we anticipate that $\pi_{\theta}$ will exhibit strong generalization on $\mathcal{D}_{\text {test}}$.

\subsection{Environmental Dynamics Modeling with Diffusion}
Since there is no public basketball environment that is able to provide online simulation, previous studies focus on offline simulations \citep{chen2022reliable}.
However, these approaches fall short in providing trajectories with planning strategies and efficiency due to the autoregressive designs, which are also challenging to be extended to incorporate dynamic planning.
Therefore, we adopt diffusion models not only to simulate trajectories simultaneously from modeling environmental dynamics but also to be guided by the specific outcomes with conditional sampling.

\noindent \textbf{Model Input and Output.}
To represent our input that can be consumed by the diffusion model, we represent all the trajectories in the format of a 2-dimensional image as described in Figure~\ref{fig:input}. To be specific, we concatenate the state features and action features at each timestep in the game together to form one column of the model input. The features from different timesteps are then stacked together following the temporal order to form the rows. In other words, the rows in the model input correspond to the \textit{planning horizon} $T$ in Section~\ref{sec:background_traj_opt}.

\noindent \textbf{Architecture.}
As illustrated in Figure~\ref{fig:architect}, the backbone of the environmental dynamics modeling module is a diffusion probabilistic model $\epsilon_\theta$. Diffusion models have been found effective in fitting the distribution of images \citep{ho2020denoising}. Our assumption is that the diffusion models can also learn the underlying distribution of basketball player trajectories by framing as the trajectory optimization problem, thereby modeling the player and ball dynamics. Following image-based diffusion models, we adopt the U-net architecture \citep{ronneberger2015u} as the overall architecture. Moreover, 
to account for the temporal dependencies between different timesteps of the trajectories, we replace two-dimensional spatial convolutions with one-dimensional temporal convolutions. 

\noindent \textbf{Diffusion Training.}
We follow the usual way by parameterizing the Gaussian noise term to make it predict $\epsilon_t$ from the input $x_t$ at diffusion step $t$ to learn the parameters $\theta$,:

\begin{equation}
\mathcal{L}(\theta) = \expect{
    t,\epsilon_t,\btau{0}
    }{
    \lVert \epsilon_t - \epsilon_\theta(\btau{t}, t) \rVert^2
    },
\end{equation}

where $\epsilon_t \sim \mathcal{N}(\bm{0}, \bm{I})$ denotes the noise target, $t$ represents the diffusion step, and $\btau{t}$ is the trajectory $\btau{0}$ corrupted by noise $\epsilon$ at diffusion step $t$.

\subsection{Value Function Training for Reward Model}
At the heart of the value function is an encoder that takes the trajectory data as input and returns the estimated cumulative reward. The structure of the return predictor $\mathcal{J}_\phi$ takes exactly the first half of the U-Net employed in the diffusion model, and it is followed by a linear layer that generates a single scalar output indicating the reward value.

\subsection{Guided Planning as Conditional Sampling}

Existing studies \citep{janner2022planning,ajay2022conditional} have revealed the connections between classifier-guided / classifier-free sampling \citep{dhariwal2021diffusion} and reinforcement learning. The sampling routine of \ours resembles the classifier-guided sampling. In detail, we condition a diffusion model $p_\theta(\btau{})$ on the states and actions encompassed within the entirety of the trajectory data. Following this, we develop an isolated model, $\mathcal{J}_\phi$, with the aim of forecasting the aggregated rewards of trajectory instances $\btau{i}$. The trajectory sampling operation is directed by the gradients of $\mathcal{J}_\phi$, which adjust the means $\mu$ of the reverse process as per the following equations:

\begin{table*}[!t]
\begin{minipage}[]{.6\linewidth}
\centering
\renewcommand{\arraystretch}{1.4}
\scalebox{.95}{
\begin{tabular}[]{cccc}
\toprule
\textbf{\# Training Games} & \textbf{\# Minutes} & \textbf{\# Plays} & \textbf{\# Frames}    \\ \midrule
480               & 23, 040    & 210, 952 & 34, 560, 000 \\ \midrule
\textbf{\# Testing Games}  & \textbf{\# Minutes} & \textbf{\# Plays} & \textbf{\# Frames}    \\ \midrule
151               & 7, 248     & 68, 701  & 10, 872, 000 \\ \midrule
\textbf{\# Games}          & \textbf{\# Minutes} & \textbf{\# Plays} & \textbf{\# Frames}    \\ \midrule
631               & 30, 288    & 279, 653 & 45, 432, 000 \\ \bottomrule
\end{tabular}}
\caption{\textbf{NBA 2015 - 16 Regular Season Game Stats}. Games are split chronically so that all the games in the test set are after any game in the training set. }
\label{tab:dataset}
\end{minipage}
\begin{minipage}[t]{.04\linewidth}~\end{minipage}
\begin{minipage}[]{.35\linewidth}
\centering
\renewcommand{\arraystretch}{1}
\begin{tabular}[]{c|c}
\textbf{Event type} & \textbf{Reward} \\ \toprule
"start of period"    & 0               \\
"jump ball"          & 0               \\
"rebound"            & 0.25            \\
"foul"               & -0.25           \\
"turnover"           & -1              \\
"timeout"            & 0               \\
"substitution"       & 0               \\
"end of period"      & 0               \\
"violation"          & -0.25           \\
"3 pointer made"     & 3               \\
"2 pointer made"     & 2               \\
"free-throw made"    & 1     
\end{tabular}
\caption{\textbf{Definition of Reward per possession.}}
\label{tab:reward}
\end{minipage}
\end{table*}

\begin{equation}
\begin{aligned}
& \mu \leftarrow \mu_\theta\left(\boldsymbol{\tau}^i\right), \\
& \boldsymbol{\tau}^{i-1} \sim \mathcal{N}\left(\mu+\alpha \Sigma \nabla \mathcal{J}_\phi(\mu), \Sigma^i\right), \\
& \boldsymbol{\tau}_{\boldsymbol{s0}}^{i-1} \leftarrow \boldsymbol{s},
\end{aligned}
\end{equation}
where $\alpha$ is the scaling factor to measure the impact of the guidance on the sampling, and
\begin{equation}
\nabla \mathcal{J}(\mu) = \left.\sum_{t=0}^T \nabla_{\mathbf{s}_t, \mathbf{a}_t} r\left(\mathbf{s}_t, \mathbf{a}_t\right)\right|_{\left(\mathbf{s}_t, \mathbf{a}_t\right)=\mu_t}.
\end{equation}
where $r$ is the reward function given by the environment. In our case, it comes from the outcome of the possessions derived from $\mathcal{D}^{pbp}$.
The detailed algorithm of reward-guided planning is illustrated in Algorithm~\ref{alg:rewardplanning}.
\begin{algorithm}
\caption{Reward Guided Planning}\label{alg:rewardplanning}
\begin{algorithmic}
\STATE \textbf{Require} diffusion model $\mu_\theta$, guide $\mathcal{J}_\phi$, scale $\alpha$, covariances $\Sigma^i$
\WHILE{not done}
    \STATE Acquire state $\bs$; initialize trajectory $\btau{N} \sim \mathcal{N}(\bm{0}, \bm{I})$ \\
    \STATE \small{\color{gray}\te{//$N$ diffusion steps in total}} \\
    \FOR{$i = N, \ldots, 1$}
        \STATE $\mu \gets \mu_\theta(\btau{i})$ \\
        \STATE $\btau{i-1} \sim \mathcal{N}(\mu + \alpha \Sigma \nabla \mathcal{J}(\mu), \Sigma^i)$\\
        \STATE \small{\color{gray}\te{//conditioned on the initial player positions}}
        \STATE $\btau{i-1}_{\vs_0} \gets \vs$ \hspace{0.15cm} \\
    \ENDFOR
    \STATE Execute first action of trajectory $\btau{0}_{\ba_0}$
\ENDWHILE
\end{algorithmic}
\end{algorithm}

\section{Experiments}
\label{sec:exp}

\subsection{Experimental Setup}
To quantitatively evaluate the effectiveness of player behavior planning, we focus on measuring the cumulative return given by the learned policy, which serves as an objective evaluation metric to compare the performance of \ours with other comparative methods. Evaluating offline RL is inherently difficult as it lacks   real-time environment interaction for reward accumulation. Thus the model verification is primarily reliant on utilizing existing replay data. To validate the capacity of our framework in learning efficient tactics, we assess \ours's ability to generate efficient plans using diverse data of varying standards.



\noindent\textbf{Dataset.}
%
We obtained our data from an open-source repository \citep{sportvu2016git,pbp2016bdb}. 
The model's input data is a combination of two components:
(1) \textbf{Player Movement Sensor Data}: This component captures real-time court events, detailing the positions of the players and the ball in Cartesian coordinates. The sampling frequency of this data is 25 frames per second. The statistics are detailed in Table~\ref{tab:dataset}.
(2) \textbf{Play-by-Play}: This segment of information contains the specifics of each possession, such as the termination of the possession (whether through a jump shot, layup, foul, and so forth), the points gained by the offensive team, the location from which the ball was shot, and the player who made the shot, among other details. 
The data for training and testing is split chronologically: the training set includes games from 2015, amounting to 480 games, while the remaining games from 2016 form the testing set, amounting to 151 games. The statistics are described in Table \ref{tab:dataset}.

\noindent\textbf{Reward Definition.}
\label{reward-definition}
As there is no fine-grained reward design in basketball in previous work, e.g., \cite{DBLP:conf/aaai/YanaiSKSR22,chen2022reliable}, we define the reward of each possession based on its outcomes, as listed in Table~\ref{tab:reward}. For a certain team that plays the possession, we encourage the possession trajectory if it leads to positive outcomes (e.g., score, rebound) and we punish otherwise (turnover, foul, violation). Note that the same event by the opponent team takes the negative value of the rewards. For example, a 2-point basket made by the team on offense leads to a $-2$ reward to the training sample of the value function for the team on defense.
During our offline evaluation, we employ our value function $\mathcal{J}_\phi$ to gauge the expected return of our policy. By summing all expected rewards from each possession for a team, we can approximate the total points for the team following the learned strategic policies. For each game in the test set, all comparative methods plan trajectories from each possession’s actual initial state.

\begin{table*}[t!]
\centering
\renewcommand{\arraystretch}{1.1}
\begin{tabular}{lcccccc}
\toprule
\textbf{Methods}                           & \textbf{Random Walk} & \textbf{Ground Truth} & \textbf{BCQ} & \textbf{CQL} & \textbf{IQL} & \textbf{\ours}        \\ \midrule
\multicolumn{1}{l|}{\textit{\textbf{AVG}}} & -9.1172±0.035        & 0.0448±0.000          & 0.0964±0.000 & 0.0986±0.001 & 0.0992±0.000 & \textbf{0.4473±1.235} \\ \midrule
\multicolumn{1}{l|}{\textit{\textbf{MAX}}} & -9.0753              & 0.0448                & 0.0967       & 0.0995       & 0.0992       & \textbf{2.2707}       \\ \bottomrule
\end{tabular}
\caption{\textbf{Overall performance in return values per possession.}}
\label{tab:main}
\end{table*}

\noindent\textbf{Baselines.}
As this task has yet to be explored, there are no widely adopted baselines for direct comparison.
Therefore, we examine our model with several state-of-the-art offline RL algorithms and a naive baseline to verify its effectiveness:

\begin{itemize}[nosep,leftmargin=*]
    \item Batch-Constrained deep Q-learning (\textbf{BCQ}) \citep{fujimoto2019off} is an off-policy algorithm for offline RL. It mitigates overestimation bias by constraining the policy to actions similar to the behavior policy, ensuring a more conservative policy.
    \item Conservative Q-Learning (\textbf{CQL}) \citep{kumar2020conservative} is an offline RL approach that minimizes an upper bound of the expected policy value to conservatively estimate the action-value function, leading to a more reliable policy.
    \item Independent Q-Learning (\textbf{IQL}) \citep{kostrikov2021offline} is a multi-agent reinforcement learning approach where each agent learns its own Q-function independently. It offers an efficient solution for multi-agent environments.
    \item \textbf{Random Walk} is the ``naive'' baseline that can be used to validate the correctness of the value function and to offer an auxiliary comparative method that corresponds to the case where all the players navigate randomly within the range of the court.
\end{itemize}


\subsection{Implementation Details}
\label{sec:imple_det}
We set the planning horizon length to $1,024$ so that all trajectories in the training data can be fitted in our diffusion model. The diffusion step is set to $20$ in all experiments. The learning rate is $2 \times 10^{-5}$ without learning rate scheduling. The hidden dimension is set following \cite{janner2022planning}. The training batch size is set to $512$. We train all models for $245K$ training steps. The value function is optimized with the mean square error loss.
All experiments are run on the NVIDIA Tesla V100 Tensor Core GPUs with 16GB memory.

\subsection{Overall Performance}
Table~\ref{tab:main} shows the cumulative scores of the generated trajectories of the compared methods. For all the models, we run each 5 times and report the average performance with the corresponding variance. 
We observe that: (1) \ours consistently and significantly outperforms the baselines and the historical gameplay in generating trajectories with higher rewards. (2) The dedicated offline RL baselines CQL and IQL are also able to learn from historical replays with mixed rewards. However, they perform noticeably worse than 
\ours, indicating that the diffusion model in \ours better captures the intrinsic dynamics of basketball gameplay. (3) As expected, the random walk baseline performs poorly, further highlighting the effectiveness of the value function in distinguishing between superior and inferior planning trajectories. These observations suggest that the diffusion model is a powerful tool of modeling complex environmental dynamics and, when combined with guided sampling, becomes a strong planning tool.


\begin{table}[t!]
\centering
\resizebox{\linewidth}{!}{
\begin{tabular}{lccccc}\toprule
$\alpha$ & \textbf{0} & \textbf{0.01} & \textbf{0.1} & \textbf{1} & \textbf{10} \\ \midrule
\multicolumn{1}{l|}{\textit{\textbf{AVG}}} &0.0859±0.0052 &0.0894±1.2263 &0.4473±1.2349 &3.0870±1.4955 & 10.8090±2.4050 \\ \midrule
\multicolumn{1}{l|}{\textit{\textbf{MAX}}} &0.0932 &1.8844 &2.2707 &5.3534 &14.2389 \\ \bottomrule
\end{tabular}
}
\caption{\textbf{The effects of the scaling factor $\alpha$.} We repeat our sampling process $5$ times and report the mean and variance for the average returns per possession.}
\label{tab:ablation}
\end{table}

\begin{figure}[t] 
\centering
\begin{subfigure}{0.3\linewidth}
  \includegraphics[width=\linewidth]{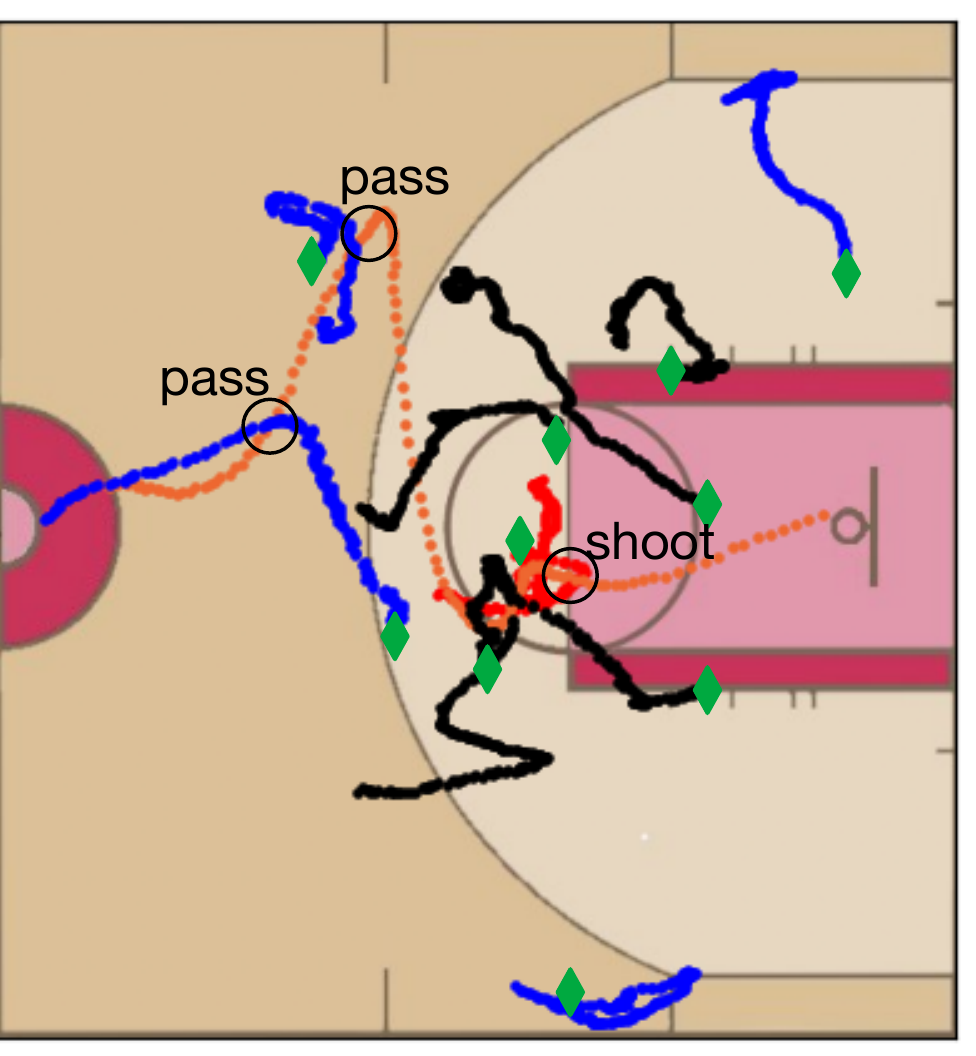}
\caption{Reward: 2.194}
\label{fig:cs_1}
\vspace{1pt}
\end{subfigure}
\ \
\begin{subfigure}{0.3\linewidth}
  \includegraphics[width=\linewidth]{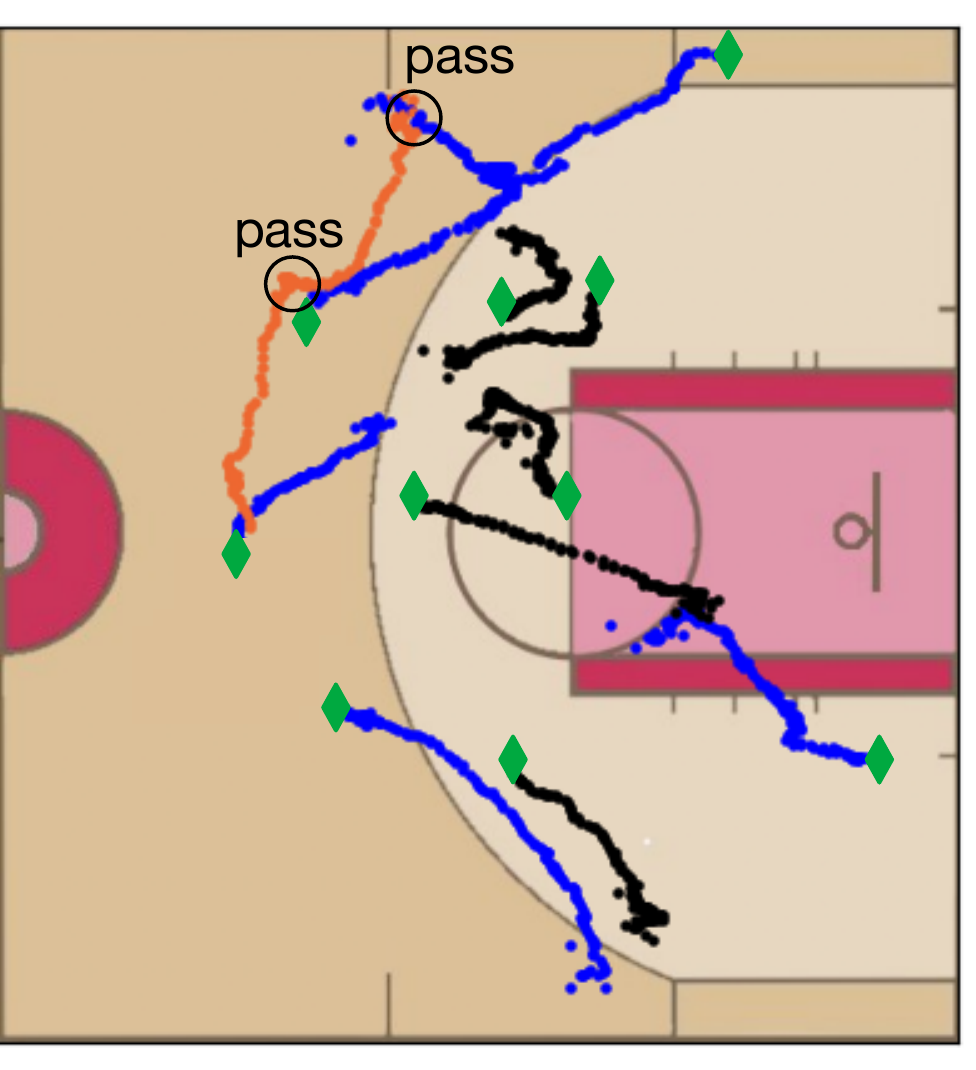}
\caption{Reward: 0.864}
\label{fig:cs_2}
\end{subfigure}
\ \
\begin{subfigure}{0.3\linewidth}
  \includegraphics[width=\linewidth]{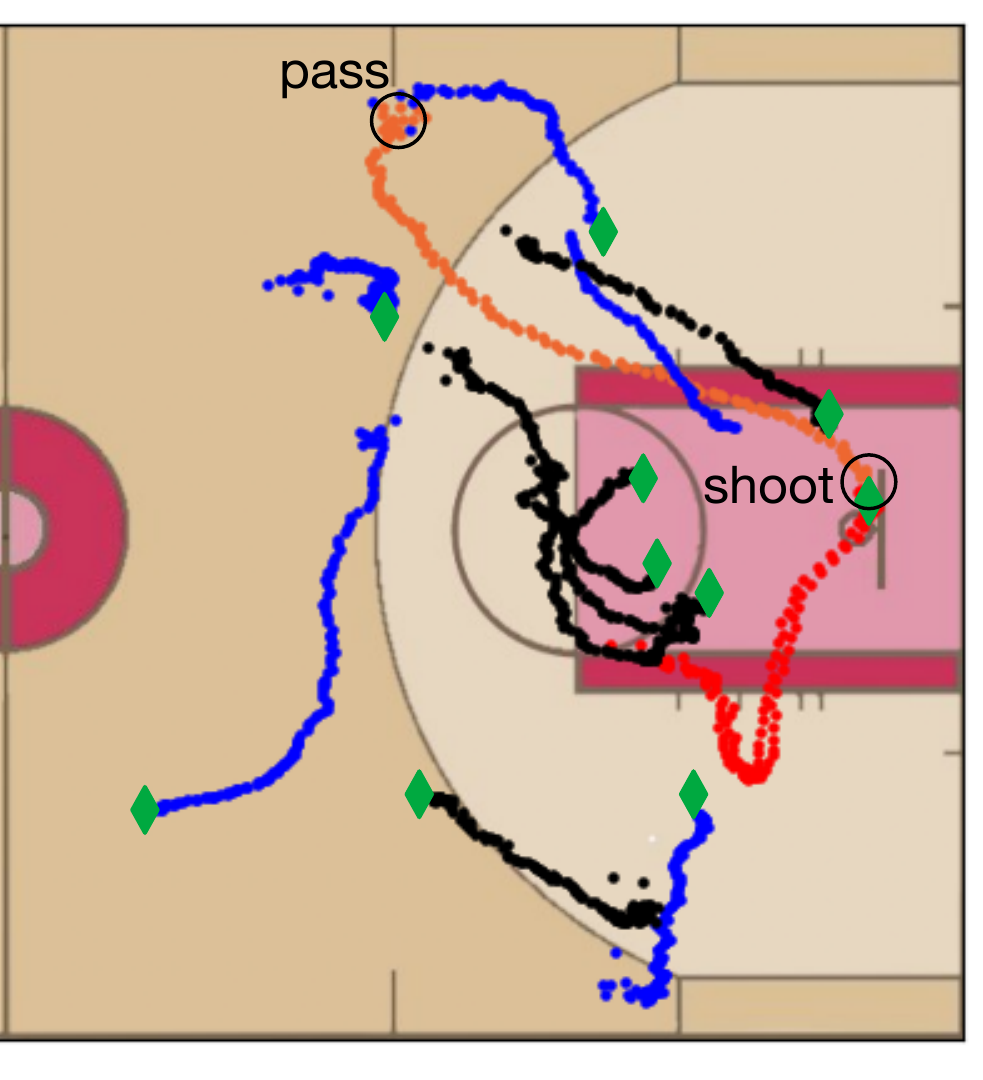}
\caption{Reward: 1.541}
\label{fig:cs_3}
\end{subfigure}
\caption{\textbf{(a, b, c)}: \textbf{Sampled cases of possessions generated by \ours.} \ours learns strategies deviating from existing data yet still aligning with subjective expectations for effective basketball play. 
The \textbf{{\color{blue} blue}} team is on offense and moves towards the right basket, while the \textbf{black} team is on defense. The ball is marked in \textbf{{\color{orange} orange}}. The player who scores for the \textbf{{\color{blue} blue}} team is highlighted in \textbf{{\color{red}Red}} (no shot attempt in (b)). 
Diamonds({\color{OliveGreen}$\blacklozenge$}) are final positions of the players. More details are in Section \ref{sec:case_study}.
}
  \label{fig:cs_1st_set}
\end{figure}

\subsection{Analysis}
\label{sec:analysis}
Table~\ref{tab:ablation} demonstrates the overall return evaluated on all the trajectories generated by \ours with $\alpha$ being $\{0, 0.01, 0.1, 1.0, 10.0\}$. It is noted that $\alpha=0$ indicates \ours performing unconditional sampling without the perturbation of the gradient of the value function.

\subsubsection{Hyperparameter Study} 
When the diffusion model performs conditional sampling for trajectories, the scaling factor $\alpha$ 
serves as a balance between quantitative scores and interpretability.
\begin{figure}[t] 
\centering
\begin{subfigure}{0.3\linewidth}
  \includegraphics[width=\linewidth]{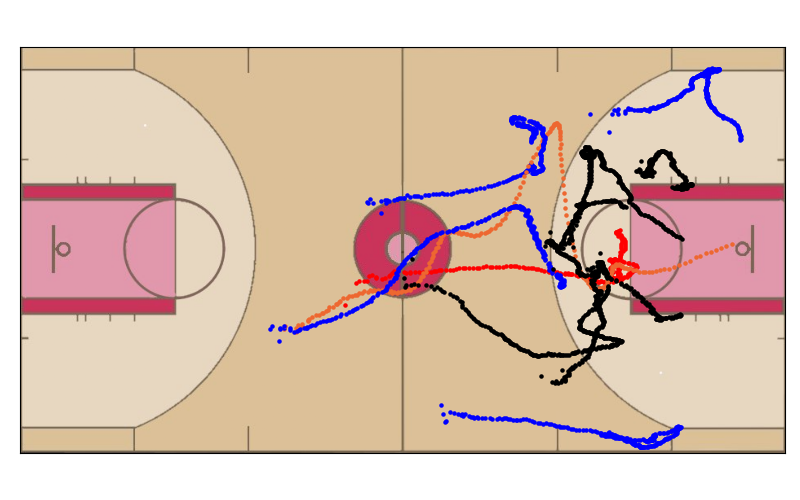}
\caption{$\alpha=0.1$}
\label{fig:alpha01}
\end{subfigure}
\ \
\begin{subfigure}{0.3\linewidth}
  \includegraphics[width=\linewidth]{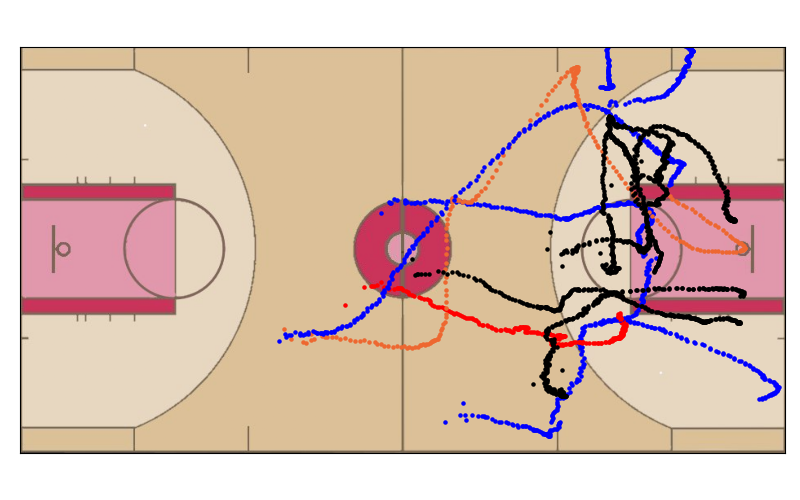}
\caption{$\alpha=1.0$}
\label{fig:alpha1}
\end{subfigure}
\ \
\begin{subfigure}{0.3\linewidth}
  \includegraphics[width=\linewidth]{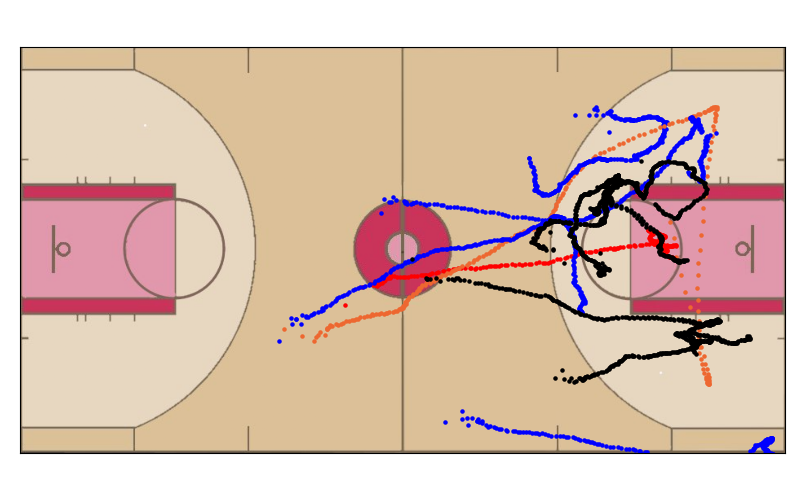}
\caption{$\alpha=10.0$}
\label{fig:alpha10}
\end{subfigure}
\caption{\textbf{(a, b, c)}: \textbf{Possessions generated by \ours with different $\alpha$.}
}
  \label{fig:alpha_study}
\vspace{-12pt}
\end{figure}
With the increase of $\alpha$, the value guidance generally has a larger impact and improves the overall cumulative rewards on the test games. 
Then the question becomes, \textit{why not keep increasing the value of $\alpha$?} To provide a deeper insight into this, we conduct a comparative study demonstrated in Figure~\ref{fig:alpha_study}. 
We consider trajectories initiated from the same state but with different scaling factors, specifically $\alpha$ values of $0.1$, $1.0$, and $10.0$. By visualizing these trajectories, we aim to demonstrate how variations in the scaling factor can significantly influence the progression and outcomes of the game, further emphasizing the crucial role of this parameter in our model.
When $\alpha=1.0$, there seems to be a mysterious force that pulls the ball to the basket. In the $\alpha=10.0$ case, the synthesized trajectory becomes even less interpretable since the ball never goes through the basket. In both $\alpha=1.0$ and $\alpha=10.0$ cases, the ball exhibits behaviors that defy the laws of physics, seemingly being propelled towards the basket as if being controlled by an invisible player.

\subsubsection{Ablation Study}
The full \ours model with sufficient value guidance outperforms the ablation version (i.e., $\alpha=0$), indicating the necessity of the value guidance. By mere unconditional sampling, the ablation version is already able to generate on average better plans than the ground truth plays in the test set. These observations confirm our two claims: The value-based guided sampling directs the diffusion model to generate trajectories leaning towards the higher-reward regions of the state-action space; and the diffusion model on its own can generate coherent and realistic trajectories representing a competent game plan.

\subsubsection{The adversary of the game}
Notably, basketball games and many other team sports are adversarial. We implemented additional defensive strategies including man-to-man and 2-3 zone defense, and ran the learned policy against these strategies \textit{iteratively} to add adversaries. In each iteration, \ours samples a trajectory of length $m$, and we replace the trajectories corresponding to \textit{defensive} players (5 channels) with those generated with man-to-man or 2-3 zone defense. The trajectories on the defensive side act as adversarial agents competing against the diffusion policy. 
The results are reported in Table~\ref{tab:def_heuristics}.
We observe that: (1) The offensive strategy encoded in \ours outplays the man-to-man defense and 2-3 zone defense. (2) When increasing the length of the segment of the trajectory, \ours is more likely to generate more coherent trajectories, leading to better returns when faced with the same defense.

\begin{table}[t]
\centering
\resizebox{\linewidth}{!}{
\begin{tabular}{lcccc}
\toprule
\textbf{length $m$} & \textbf{25} & \textbf{50} & \textbf{75} & \textbf{100} \\
\midrule
\textbf{man-to-man} & $1.410 \pm 0.368$ & $1.750 \pm 0.059$ & $2.526 \pm 0.039$ & $2.814 \pm 0.008$ \\ \midrule
\textbf{2-3 zone} & $1.424 \pm 0.284$ & $1.558 \pm 0.309$ & $2.229 \pm 0.011$ & $2.327 \pm 0.029$ \\
\bottomrule
\end{tabular}}
\caption{Return values competing against defense.}
\label{tab:def_heuristics}
\end{table}

\subsection{Case Study}
\label{sec:case_study}
We now perform a case study to qualitatively demonstrate the practicability of value-guided conditional generation. Figure~\ref{fig:cs_1st_set} shows three cases, all of which are sampled from the trajectories generated by \ours. 
In Figure~\ref{fig:cs_1}, we visualize a possession generated with a high reward. The players in the {\color{blue} blue} team share the ball well and managed to find the {\color{red}red} player near the free-throw line. At the time the {\color{red}red} player shoots the ball, no defender is between him and the basket. The outcome of this simulated play is a 2-point basket. 
In Figures~\ref{fig:cs_2} and \ref{fig:cs_3}, two different plans with the same horizon are generated by \ours given the same initial player and ball positions. In Figure~\ref{fig:cs_2}, we observe a more conservative strategy where the ball is repeatedly passed between the {\color{blue} blue} players near the perimeter, which is also valued with a lower reward. In spite of the same initial conditions, \ours generates a more aggressive strategy in Figure~\ref{fig:cs_3} in that the ball is passed directly to the low post that leads to a 2-point basket, suggesting an aggressive tactic execution.
These cases illustrate that \ours is able to not only synthesize realistic trajectories but also output high-reward and diverse trajectories for planning tactics as well as for enhancing decision-making. 

\section{Related Work}
\label{sec:related}
\noindent\textbf{Reinforcement Learning for Planning}.
Reinforcement learning is a learning-based control approach. A wide range of application domains have seen remarkable achievements through the use of reinforcement learning algorithms, such as robotics \citep{kalashnikov2018qt}, autonomous vehicles \citep{balaji2019deepracer}, industrial regulation \citep{gasparik2018safety}, financial sectors \citep{meng2019reinforcement}, healthcare \citep{yu2019reinforcement}, 
gaming \citep{silver2017mastering}, and marketing \citep{jin2018real}. Despite its wide use, many RL applications depend on an online environment that facilitates interactions. In numerous circumstances, acquiring data online is either expensive, unethical, or dangerous, making it a luxury. Consequently, it is preferable to learn effective behavior strategies using only pre-existing data. Offline RL has been suggested to fully utilize previously gathered data without the need for environmental interaction \citep{fujimoto2019off,agarwal2020optimistic,kumar2020conservative,levine2020offline,fu2020d4rl}, which has found applications in areas such as dialogue systems \citep{jaques2019way}, robotic manipulation techniques \citep{kalashnikov2018scalable}, and navigation \citep{kahn2021badgr}.

\noindent\textbf{Sports \& Machine Learning}.
Machine learning and AI have recently been employed in sports analytics to comprehend and advise human decision-making \citep{aoki2017luck,ruiz2017leicester,decroos2018automatic,sun2020cracking,tuyls2021game,robberechts2021bayesian,wang2022stroke}.
\cite{luo2021inverse} suggested a player ranking technique that combines inverse RL and Q-learning.
\cite{wang2022stroke} proposed a deep-learning model composed of a novel short-term extractor and a long-term encoder for capturing a shot-by-shot sequence. 
\cite{wang2022shuttlenet} developed a position-aware fusion framework for objectively forecasting stroke returns based on rally progress and player style.
\cite{chang2022will} predicted returning strokes and player movements based on previous strokes using a dynamic graph and hierarchical fusion approach. While these methods are effective for producing simulations, they may not fully address the goal of maximizing specific objectives (e.g., winning games).
Previous basketball analytics mainly focused on employing recurrent neural networks to analyze player-tracking data for offensive tactics identification and player movement prediction  \citep{mcintyre2016recognizing,wang2016classifying,tian2020use,terner2020modeling}. However, these methods lack labeled interactions between the learning agent and the environment, limiting their ability to uncover optimal decision sequences.
Wang et al.~\cite{wang2018advantage} explored the use of RL to improve defensive team decisions, especially  the execution of a "double team" strategy. 
Liu et al.~\cite{liu2018learning} designed a method using motion capture data to learn robust basketball dribbling maneuvers by training on both locomotion and arm control, achieving robust performance in various scenarios.

\vspace{-5pt}

\section{Conclusion}
\label{sec:concl}

In this paper, we introduce \ours, the diffusion model with conditional sampling in planning high-rewarded basketball trajectories and synthesizing adaptive play strategies.
With the extension of environmental dynamics into the diffusion model and fine-grained rewards for the value function, \ours has shown impressive capabilities in capturing the intricate dynamics of basketball games and generating play strategies that are consistent with or even surpass professional tactics. Its adaptive nature has allowed for swift adjustments to evolving conditions and facilitated real-time identification of optimal solutions.
Extensive simulation studies and analysis of real-world NBA data have confirmed the advantages of \ours over traditional planning methods. The generated trajectories and play strategies not only outperform conventional techniques but also exhibit a high level of alignment with professional basketball tactics.
Future work will explore the integration of additional sources of information, such as player fatigue and skill levels, into our framework to further enhance its performance.
In addition, we plan to develop an open environment and a set of benchmarks to not only facilitate research on machine learning for sports but also extend to other real-time dynamic systems.
\newpage
\balance
\bibliographystyle{ACM-Reference-Format}
\bibliography{cikm24}

\end{document}